\documentclass{Interspeech2024}
\usepackage{multirow}
\usepackage{multicol}

\interspeechcameraready

\title{2DP-2MRC: 2-Dimensional Pointer-based Machine Reading Comprehension Method for Multimodal Moment Retrieval}

\name[affiliation={1}]{Jiajun}{He}
\name[affiliation={2}]{Tomoki}{Toda}

\address{
  $^1$Graduate School of Informatics, Nagoya University, Japan\\
  $^2$ Information Technology Center, Nagoya University, Japan }
\email{jiajun.he@g.sp.m.is.nagoya-u.ac.jp, tomoki@icts.nagoya-u.ac.jp}

\keywords{moment retrieval, 2-dimensional pointer, machine reading comprehension}

\begin{document}

\maketitle

\begin{abstract}
Moment retrieval aims to locate the most relevant moment in an untrimmed video based on a given natural language query. Existing solutions can be roughly categorized into moment-based and clip-based methods. The former often involves heavy computations, while the latter, due to overlooking coarse-grained information, typically underperforms compared to moment-based models. Hence, this paper proposes a novel \textbf{2-D}imensional \textbf{P}ointer-based \textbf{M}achine \textbf{R}eading \textbf{C}omprehension for \textbf{M}oment \textbf{R}etrieval \textbf{C}hoice (2DP-2MRC) model to address the issue of imprecise localization in clip-based methods while maintaining lower computational complexity than moment-based methods. Specifically, we introduce an AV-Encoder to capture coarse-grained information at moment and video levels. Additionally, a 2D pointer encoder module is introduced to further enhance boundary detection for target moment. Extensive experiments on the HiREST dataset demonstrate that 2DP-2MRC significantly outperforms existing baseline models.
\end{abstract}

\vspace{-2mm}
\section{Introduction}
\begin{figure}[t]
  \centering
    \includegraphics[scale=0.295]{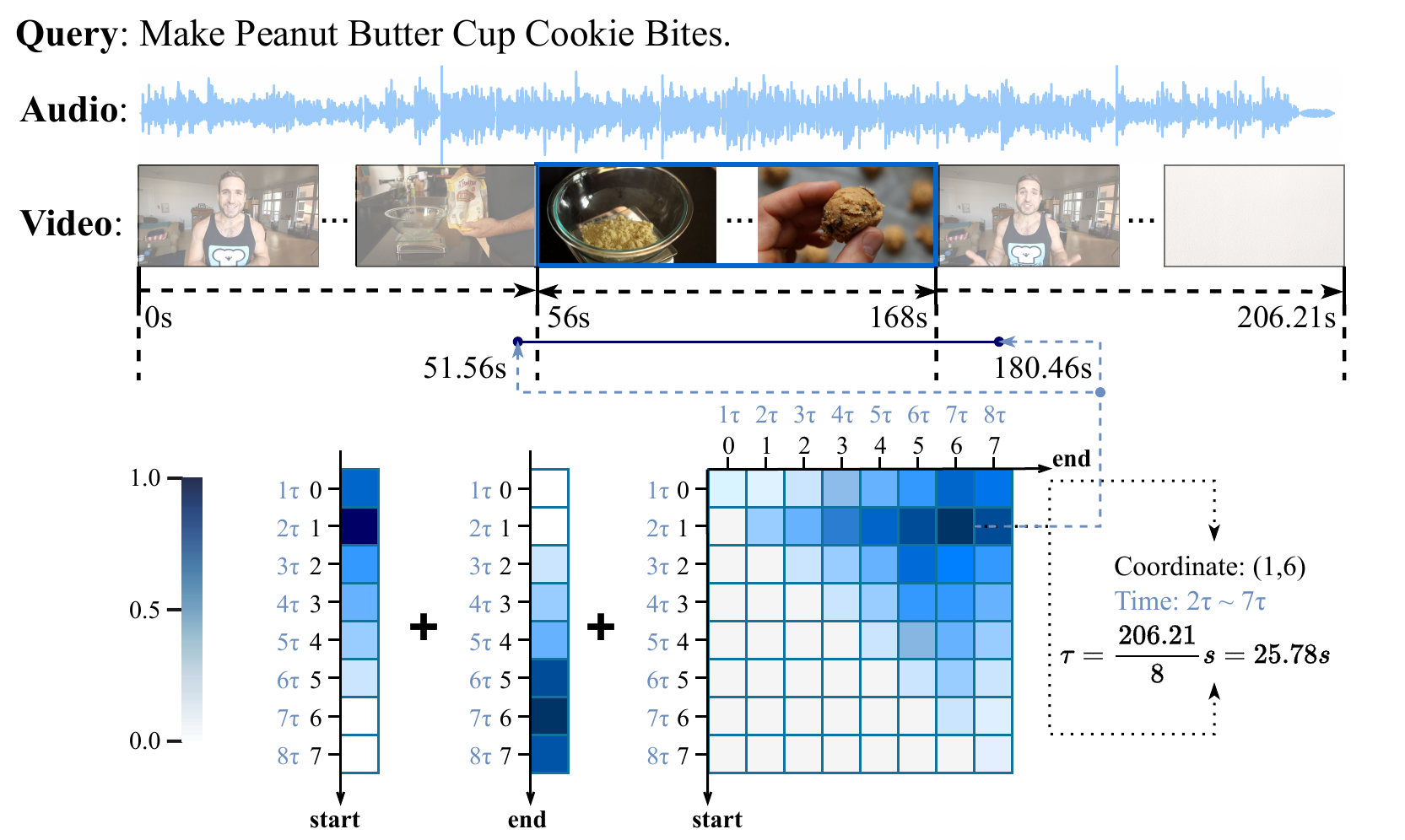}
    \caption{An example of our multimodal moment retrieval with a query in an untrimmed video. The most relevant moment is retrieved by two 1D probability matrices and a 2D probability matrix together. Note that the length of the video and the sampling rate determine the value of the short duration $\tau$ and the precision of the retrieved moments.}
  \vspace{-6mm}
  \label{fig:pipeline}
\end{figure}

\vspace{-1mm}
With the explosive growth of online video content, there is an increasingly urgent demand for rapidly and accurately retrieving the desired information from massive videos \cite{chen2024vast, huo2023weakly, zhao2023diffusionvmr}. To address this need, Hendricks et al. \cite{anne2017localizing} and Gao et al. \cite{gao2017tall} leveraged annotated language data to introduce the video moment retrieval (VMR) task. Specifically, given a natural language query and a corresponding untrimmed video, this task aims to determine the start and end time boundaries of moments that correspond to the given description \cite{liu2023survey, luo2024zero, chen2023cross}, as illustrated in Fig. \ref{fig:pipeline}. 
However, due to the need for aligning complex video (and potentially audio) content with query context and understanding the semantics between different modalities, moment retrieval becomes a meaningful yet challenging task \cite{chen2024multilevel, zhang2022dual}.

The current VMR methods can be divided into moment-based methods and clip-based methods based on whether predefined candidate moments need to be generated \cite{zhang2022video}. As clips denote the briefest moments, the contextual information at the clip level is inherently a subset of the context at the moment level.
Most moment-based approaches follow a ``propose-then-rank" process \cite{anne2017localizing, gao2017tall, zhang2022dual, li2024momentdiff, sun2023video}. 
These methods usually first generate a series of predefined candidate moments, then compare each candidate moment with the query, and finally select the best moment through ranking. Although current moment-based approaches are intuitive and achieve some good results, they suffer from issues of high computational complexity in densely generating candidate moments. 
Clip-based approaches directly align video clips with queries and predict their matching scores without generating candidate moments, thus reducing the computational cost \cite{yu2024self, huo2023semantic, wang2023spatiotemporal, tan2023vmlh}. 
However, the current methods focus on aligning clip-level video features with queries, overlooking coarse-grained moment-level and video-level information, leading to imprecise localization and performance issues.  

\begin{figure*}[htbp]
  \centering
  \includegraphics[scale=0.24]{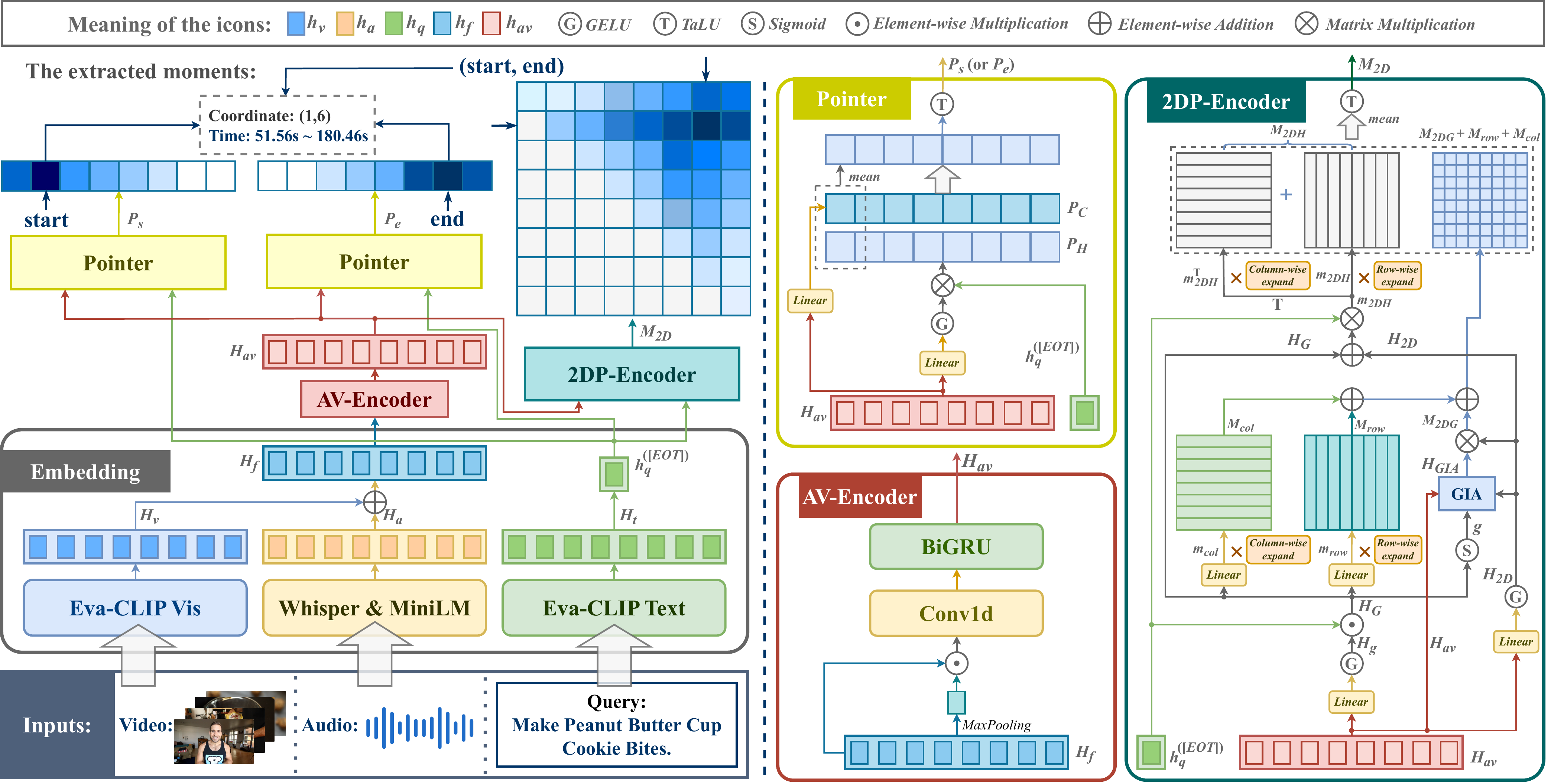}
  
  \caption{Overall architecture of the proposed 2DP-2MRC model.}
  \vspace{-4mm}
  \label{fig:model}
\end{figure*}

To address the aforementioned issues of high computational complexity and imprecise localization, we propose a novel clip-based method, \textbf{2}-\textbf{D}imensional \textbf{P}ointer-based \textbf{M}achine \textbf{R}eading \textbf{C}omprehension for \textbf{M}oment \textbf{R}etrieval \textbf{C}hoice (2DP-2MRC) model, which aligns more closely with habits of human reading comprehension. Specifically, when reading, individuals typically start by skimming through the text passage and questions to gain a preliminary understanding. They then go back to the text, focusing on sections relevant to the questions, and thoroughly integrate the contextual information with the questions to enhance comprehension before identifying potential answers \cite{zhang2021retrospective, zheng2019human, sun2022you, jiang2021new}. Our model enhances both coarse-grained and fine-grained understanding and localization capabilities of video content by utilizing the proposed Audio-Video Encoder (AV-Encoder), Pointer, Gated Interactive Attention (GIA) mechanism, and 2D Probability Encoder (2DP-Encoder) modules. The main contributions of this paper are as follows:

\begin{itemize}[leftmargin=*]
\setlength{\topsep}{0pt}
\setlength{\itemsep}{0pt}
\setlength{\parsep}{0pt}
\setlength{\parskip}{0pt}
\item We introduce an AV-Encoder to capture coarse-grained information at the moment and video levels, providing the model with a comprehensive understanding of the video.
\item We propose a pointer module based on multi-level interactive attention to obtain preliminary prediction results of the start (or end) positions of the target moments.
\item We introduce a 2DP-Encoder module based on the GIA mechanism for precise answer localization. Specifically, we first utilize an adaptive gating unit to adjust the importance of each video clip to the answer. Next, we conduct multi-level interactive attention calculation on the query and contextual information of the video, enhancing overall attention to the target moment. By constructing a 2D position coordinate matrix for potential answers, we refine boundary detection for all possible moments, thereby enhancing extraction recall and mitigating discrepancies in disparate 1D distributions, ultimately improving retrieval accuracy.
\item We propose the novel 2DP-2MRC model building upon the AV-Encoder, Pointer, and 2DP-Encoder modules. This model utilizes parallel computing to circumvent a substantial increase in time complexity, facilitating efficient and accurate extraction of target moments from videos. Experimental results demonstrate that the 2D-2MRC model significantly outperforms existing baseline models on the HIREST dataset.

\vspace{-1mm}
\end{itemize}

\section{Proposed Method}

\subsection{Problem Formulation}
\label{section2.1}
The multimodal moment retrieval task we address can be expressed as the function $f(V, A, Q) = (t^s, t^e)$, where the video modality $V = (v_1, v_2, \cdots , v_m)$ represents a video consisting of \textit{m} clips, the audio modality $A = (a_1, a_2, \cdots , a_n)$ contains the audio extracted from the video, consisting of \textit{n} frames, and the text modality $Q = (q_1, q_2, \cdots , q_l)$ represents the query content, consisting of \textit{l} words. The goal of the multimodal moment retrieval task is to combine the video \textit{V} and audio \textit{A} to retrieve the temporal moment with a start and end time point $(t^s, t^e)$ from the video that is most relevant to the query \textit{Q}.

\vspace{-2mm}
\subsection{Embedding Module}
\label{section2.2}
\vspace{-1mm}

Following the previous work \cite{zala2023hierarchical}, our embedding module is built on three existing pretrained models: EVA-CLIP \cite{fang2023eva}, Whisper \cite{radford2023robust}, MiniLM \cite{reimers2019sentence}, as illustrated in Fig. \ref{fig:model}. These pretrained models are frozen during the training stage.

\begin{itemize}[leftmargin=*]
\setlength{\topsep}{0pt}
\setlength{\itemsep}{0pt}
\setlength{\parsep}{0pt}
\setlength{\parskip}{0pt}
\vspace{-1mm}
\item \textbf{Video Clip Representations.} Assuming the video consists of $o$ continuous sequence frames, we divide the video into multiple continuous short video clips based on every $p$ frames, and then sample $m$ video clips at fixed intervals. The visual encoder of EVA-CLIP maps \textit{m} video clips into a visual embedding, denoted as $H_v = (h_v^{(1)}, h_v^{(2)}, \cdots , h_v^{(m)}) \in \mathbb{R}^{m \times d}$, where $d$ is the the dimension of hidden representations.
\item \textbf{Text Query Representations.} The text encoder of EVA-CLIP maps text queries into a query embedding $H_q = (h_q^{(1)}, h_q^{(2)}, \cdots , h_q^{(l)}) \in \mathbb{R}^{l \times d}$, and utilizes a end-of-text token ([\textit{EOT}]) to represent the semantic information of the entire sentence, denoted by $h_q^{([EOT])} \in \mathbb{R}^{d}$.
\item \textbf{ASR Representations.} Whisper extracts speech transcriptions from audio, and the MiniLM text encoder maps the speech transcriptions into an ASR embedding $H_a = (h_a^{(1)}, h_a^{(2)}, \cdots , h_a^{(m)}) \in \mathbb{R}^{m \times d}$ by uniformly samples or pads the ASR embedding to achieve a fixed length \textit{m}, which corresponds to the number of video clips.
\item \textbf{Fused Representations.} Following the previous work \cite{zala2023hierarchical}, we fuse the visual embedding and the ASR embedding using element-wise addition to obtain the fused embedding, denoted as $H_f = (h_f^{(1)}, h_f^{(2)}, \cdots , h_f^{(m)}) \in \mathbb{R}^{m \times d}$.
\end{itemize}

\vspace{-2mm}
\subsection{AV-Encoder Module}
\vspace{-1mm}
To acquire the coarse-grained feature of the fused representations, we begin by employing $\text{MaxPooling}$ across clip-wise features and then merge them with the clip-wise features. Subsequently, we utilize a temporal 1D convolutional layer \cite{kiranyaz20211d} followed by a bidirectional GRU layer \cite{jozefowicz2015empirical} to capture the local dependencies among video clips and the sequential characteristics of the video, represented as $H_{av} = (h_{av}^{(1)}, h_{av}^{(2)}, \cdots , h_{av}^{(m)}) $:
\vspace{-2mm}
\begin{equation}
\label{eq13}
  H_{av} = \text{BiGRU}(\text{Conv1d}(\text{MaxPooling}(H_{f}) \odot H_{f})) \in \mathbb{R}^{m \times d},
  \vspace{-1mm}
\end{equation}
where $\odot$ denotes element-wise multiplication.

\vspace{-2mm}
\subsection{Pointer Module}
\vspace{-1mm}
The pointer module is responsible for mapping the fused representation of audio-integrated video clips $H_{av}$, to a probability distribution vector of length $m$, to obtain the preliminary prediction results of the start (or end) positions of the target moments. Unlike the simple linear mappings in \cite{zala2023hierarchical}, the pointer module utilizes the global query representation $h_q^{([EOT])}$ to further calculates the multi-level interactive attention between the query, the fused representation of audio-integrated video clips, and the contextual information to enhance the overall focus on the target moments.

The input representation $H_{av}$ is first fed into a linear layer, where the activation function is GELU \cite{hendrycks2016gaussian}, to obtain a new representation $H$:
\vspace{-2mm}
\begin{equation}
\label{eq13}
  H = \text{GELU}( \text {Linear}(H_{av})) \in \mathbb{R}^{m \times d}.
  \vspace{-2mm}
\end{equation}

Then $H$ interacts with the global semantic query representations $h_q^{([EOT])}$ to get the candidate probability distribution $P_H$:
\vspace{-1mm}
\begin{equation}
\label{eq13}
  P_H = \text{TaLU}( (H \cdot h_q^{([EOT])}))\in \mathbb{R}^{m},
  \vspace{-1mm}
\end{equation}
where \text{TaLU} is a normalized probabilistic function used to map the calculated interactive attention scores to the range of (0, 1). Since the derivative range of the \text{TaLU} function is four times that of the widely used Sigmoid function, it effectively enhances the differentiation of probabilities at different positions, thereby aiding in the model's boundary detection.
\vspace{-2mm}
\begin{equation}
\label{eq13}
  \text{TaLU}(x) = \frac{e^{x} }{e^{x} +e^{-x} } \in (0,1).
  \vspace{-2mm}
\end{equation}

In addition, the input representation $H_{av}$ also passes through a layer of linear compression with an output dimension of 1, to obtain another candidate probability distribution $P_C$:
\vspace{-2mm}
\begin{equation}
\label{eq13}
  P_C = \text{TaLU}( \text {Linear}(H_{av})) \in \mathbb{R}^{m}.
  \vspace{-2mm}
\end{equation}

The preliminary probability distribution $P_s$ (or $P_e$) of the start (or end) position is the mean of $P_H$ and $P_C$.

\subsection{2DP-Encoder Module}
The 2DP-Encoder enhances boundary detection and improves moment retrieval recall by mapping the input representation $H_{av}$ into a 2D probability distribution matrix with both length and width of $m$.
On one hand, the input $H_{av}$ passes through two independent linear transformations, resulting in two representations $H_{g}$ and $H_{2D}$, which are used to compute the gating matrix and the 2-dimensional probability distribution matrix:
\begin{equation}
\label{eq13}
  H_{g} = \text{GELU}( \text {Linear}(H_{av})) \in \mathbb{R}^{m \times d},
\end{equation}
\begin{equation}
\label{eq13}
  H_{2D} = \text{GELU}( \text {Linear}(H_{av})) \in \mathbb{R}^{m \times d}.
\end{equation}
%

Next, $H_{g}$ interacts with the global query representation $h_q^{([EOT])}$ to obtain representations $H_{G}$, and is normalized through Sigmoid function to the range (0,1) to derive the ultimate gating weight $g$:

\begin{equation}
\label{eq13}
\vspace{-2mm}
  H_{G} = H_{g} \odot h_q^{([EOT])} \in \mathbb{R}^{m \times d},
\end{equation}
%
\vspace{-2mm}
%
\begin{equation}
\label{eq13}
  g = \text{Sigmoid}(H_{G}) \in \mathbb{R}^{m \times d},
\end{equation}
%
where the gating weight $g$ is used to adaptively adjust the importance of the video clips related to the target moments, and the adjusted result is represented as $H_{GIA}$:
\vspace{-1mm}
\begin{equation}
\label{eq13}
  H_{GIA} = g \odot H_{2D}  + (1-g) \odot H_{av} \in \mathbb{R}^{m \times d}.
  \vspace{-2mm}
\end{equation}

Then, we compute the interactive attention between $H_{GIA}$ and $H_{2D}$, resulting in the first candidate 2D probability distribution matrix $M_{2DG}$ for the 2D position coordinates:
\vspace{-1mm}
\begin{equation}
\label{eq13}
  M_{2DG} = \text{TaLU}(H_{2D} \cdot H_{GIA}^\text{T}) \in \mathbb{R}^{m \times m},
  \vspace{-2mm}
\end{equation}
where $\text{T}$ represents the transpose operation. 

On the other hand, $h_q^{([EOT])}$ also interacts with $H_{G}$ and $H_{2D}$, and the resulting representations $m_{2DH}$ are expanded by rows and columns and then added to generate the second candidate 2D probability distribution $M_{2DH}$:
\vspace{-1mm}
\begin{equation}
\begin{split}
\label{eq13}
  m_{2DH} = \text{TaLU}((H_{G}+H_{2D}) \cdot h_q^{([EOT])}) \in \mathbb{R}^{m} \\
  \to M_{2DH}\in \mathbb{R}^{m \times m}.
\end{split}
\vspace{-2mm}
\end{equation}

To further enrich the information in the 2D probability distribution and retrieve more accurate moments, the 2DP-Encoder also considers the distribution information $M_{row}$ and $M_{col}$ expanded along rows and columns, respectively:
\begin{equation}
\label{eq13}
  m_{row} = \text{TaLU}(\text{Linear}(H_{G}^{\text{T}})) \in \mathbb{R}^{m} \\
  \to M_{row}\in \mathbb{R}^{m \times m},
\end{equation}
\begin{equation}
\label{eq13}
  m_{col} = \text{TaLU}(\text{Linear}(H_{G})) \in \mathbb{R}^{m} \\
  \to M_{col}\in \mathbb{R}^{m \times m}.
\end{equation}
%

The final output 2D probability distribution $M_{2D}$ is the mean of $M_{2DG}$, $M_{2DH}$, $M_{row}$ and $M_{col}$.

\begin{figure}[t]
  \centering
    \includegraphics[scale=0.26]{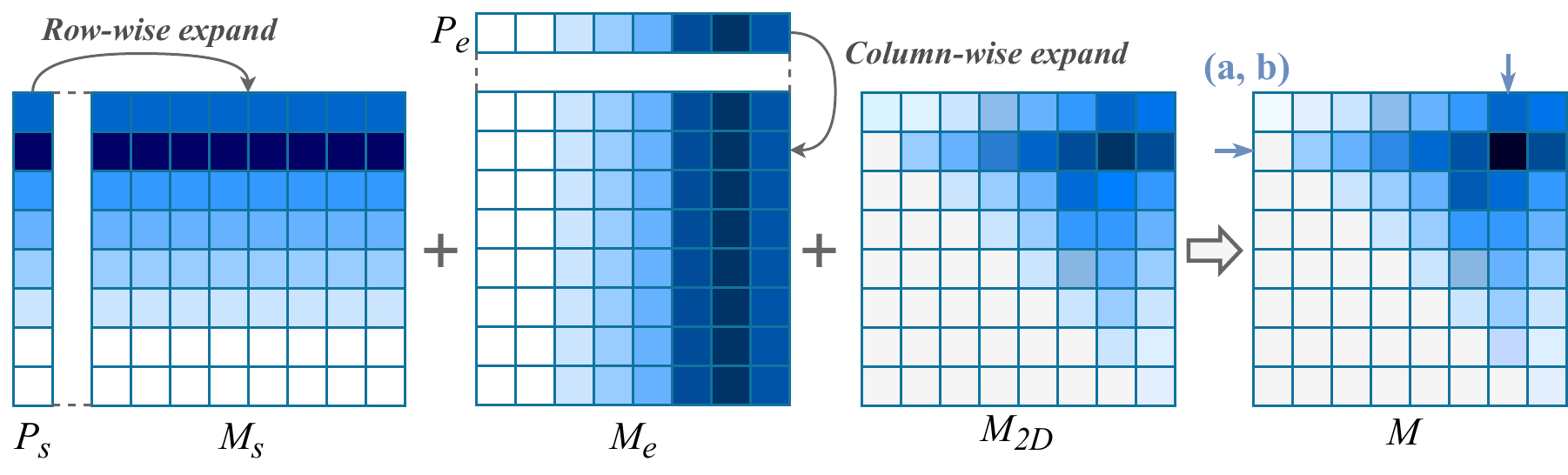}
  \caption{The process of score prediction.}
  \vspace{-6mm}
  \label{fig:score}
\end{figure}


\subsection{Score Prediction}
During scoring, we expand $P_s$ and $P_e$ into $M_s$ and $M_e$ by row and column, respectively. The final 2D score map $M$ is formed by adding $M_s$, $M_e$, and $M_{2D}$ (see Fig. \ref{fig:score}). The coordinates $(a,b)$ in $M$ represent a moment starting from clip $a$ to $b$. To maintain validity, the start and end clip indexes $a$ and $b$ must satisfy $a \leq b$. Thus, we set all elements in the lower triangle of $M$ to zero and choose the highest score as the retrieved moment.

\subsection{Loss Function}
The training loss function comprises three weighted loss terms: the loss between the predicted start position $P_s$ by the Pointer module and the target start position distribution $Y_s$, the loss between the predicted end position $P_e$ by the Pointer module and the target end position distribution $Y_e$, and the loss between the predicted 2D probability distribution $M_{2D}$ by the 2DP-Encoder module and the target position coordinate distribution $Y_m$. Here, $Y_s$, $Y_e$, and $Y_m$ are one-hot labels.
\vspace{-2mm}
\begin{equation}
\label{eq13}
\text { Loss }=\frac{1-\lambda}{2}\left[f\left(P_s, Y_{s}\right)+f\left(P_e, Y_{e}\right)\right]+\lambda \cdot f\left(M_{2D}, Y_{m}\right),
\end{equation}
where $f\left(x, y\right)$ represents the Binary Cross Entropy
(BCE) loss between predictions and ground-truth labels and $\lambda$ is a hyperparameter to adjust the weight of 2D probability distribution.

\section{Experiments}

\subsection{Experiment Settings and Evaluation Metrics}
\label{ssec:Experiment Settings}
Our method was implemented using Python 3.10.0 and Pytorch 1.11.0. The model was trained and evaluated on a computer with Intel(R) Xeon(R) Gold 6248 CPU @ 2.50GHz, 32GB RAM and one NVIDIA Tesla V100 GPU. The model was trained with the AdamW \cite{loshchilov2017decoupled} optimizer with a batch size of 16, and the weight decay of 0.01.
During training, the dropout rate was 0.3 and the weight $\lambda$ of 2D probability distribution was 0.1. In addition, we set the initial learning rate at $1e^{-3}$. The size of all hidden states in the model was set to 512. 

Following previous work \cite{iashin2020better, sun2022you, zala2023hierarchical}, we evaluated model output against the ground-truth moment with Recall@1 with Intersection over Union (IoU) thresholds (0.5 and 0.7).

\subsection{Datasets}

To evaluate the effectiveness of our proposed model, we carried out experiments on the HIREST dataset \cite{zala2023hierarchical}, tailored for various tasks including video retrieval, moment retrieval, moment segmentation, and step captioning. 
In this paper, we only focused on the moment retrieval task. HIREST consisted of a total of 3.4K text-video pairs, which were 287 seconds long on average, with a total duration of 270 hours. Out of 3.4K videos, 1.8K videos were clippable to a moment. The average moment length was 148 seconds, which was 55\% of the original videos. Due to instances of multiple videos being retrieved for a single query, HIREST was divided into train/val/test splits based on queries rather than videos, resulting in splits of 546/292/546 queries (1507/477/1391 videos) for train/val/test, respectively.

\subsection{Results and Analysis}
We compare our proposed model with three baseline models:
\begin{itemize}[leftmargin=*]
\setlength{\topsep}{0pt}
\setlength{\itemsep}{0pt}
\setlength{\parsep}{0pt}
\setlength{\parskip}{0pt}
\item The BMT event proposal module \cite{iashin2020better} is a moment-based dense video captioning model pretrained on ActivityNet Captions, which predicts video event proposals with center/length/confidence values. BMT generates various events and then take the minimum start time and maximum end time across the events as the retrieved moment.
\item The Joint model \cite{zala2023hierarchical} is a clip-based text question answering model that utilizes a multimodal encoder and two linear layers to predict the start and end positions of moments.
\item The MPGN model \cite{sun2022you} tackles moment retrieval task as a moment-based multi-choice reading comprehension task and achieves advanced performance by introducing a fine-grained feature encoder and a conditioned interaction module, etc.
\end{itemize}

Table \ref{tab:Experimental_results} presents the results of our model compared to other baselines. Overall, our method outperforms others by a large margin on the HIREST dataset. 
Furthermore, all models exhibit significant performance drops when audio input is not provided, indicating the helpfulness of audio for boundary detection.

  


    

    


\vspace{-2mm}
\begin{table}[h!]
    \caption{Experimental results. 1fps means that the model extracts one frame per second as a candidate moment.}
    \vspace{-2mm}
  \label{tab:Experimental_results}
  
  \centering
  \resizebox{\linewidth}{!}
  {
  \begin{tabular}{@{}c|ccccc@{}}
    \toprule
    \multirow{2}{*}{\textbf{Model}} & \multirow{2}{*}{{\textbf{\# of Video Clips}}} & \multicolumn{2}{c}{\textbf{Recall@1}}  & \multicolumn{2}{c}{\textbf{Recall@1}}  \\
    &  &0.5  & 0.7 &0.5  & 0.7\\
    \midrule
     & &   \multicolumn{2}{c}{\textbf{With Audio}} & \multicolumn{2}{c}{\textbf{Without Audio}}    \\
    \midrule
    \multirow{1}{*}{BMT}  & 1fps & 71.91& 39.18  & 62.6 & 32.34 \\

    \multirow{1}{*}{Joint}   & 1fps & 73.32 & 32.60 & 70.7& 20.6 \\

   \midrule
    \multirow{1}{*}{Joint}  & 64 & 73.58 & 32.12 & 71.50 &26.94  \\
    \multirow{1}{*}{MGPN}   & 64 & 74.09 & 43.01 & 73.34& 38.24 \\
    \multirow{1}{*}{\textbf{2DP-2MRC (ours)}}   & \textbf{64} & \textbf{75.13} & \textbf{54.92}  &\textbf{74.23} &\textbf{46.11} \\
    
   \midrule
    \multirow{1}{*}{Joint}   & 128 & 70.98 & 35.23 & 69.43&24.35  \\
    \multirow{1}{*}{MGPN}   & 128 & 78.76 & 55.44 & 74.09 & 43.01  \\
    \multirow{1}{*}{\textbf{2DP-2MRC (ours)}}   & \textbf{128} & \textbf{79.79} & \textbf{62.69}& \textbf{74.61}& \textbf{55.96}  \\
    \bottomrule
  \end{tabular}
  }

\end{table}
\vspace{-2mm}

\textbf{Ablation Study}. To evaluate the effectiveness of each module in our 2DP-2MRC, we conduct in-depth ablation studies as shown in Table \ref{tab:Ablation_study}.
Model 1, without the AV-Encoder module, exhibits a significant performance drop, indicating the importance of coarse-grained feature encoding.
Model 2, without the Pointer module, also shows decreased performance, highlighting the necessity of fine-grained interaction between queries and video clips.
Model 3, without the 2DP-Encoder module, demonstrates the module's role in enhancing boundary accuracy and improving fine-grained alignment between video clips and queries.
Overall, all components proposed in 2DP-2MRC significantly improve the overall performance, affirming the meaningfulness and effectiveness of the integrated reading strategy in our framework, which involves initial reading of text segments and questions, refocusing on text segments, and combining information to find answers.


\vspace{-2mm}
\begin{table}[h!]
    \caption{Effectiveness of different modules in our 2DP-2MRC.}
    \vspace{-2mm}
  \label{tab:Ablation_study}
  
  \centering
  \resizebox{\linewidth}{!}
  {
  \begin{tabular}{@{}c|ccccc@{}}
    \toprule
    \multirow{2}{*}{\textbf{Model}} & \multicolumn{3}{c}{\textbf{Modules}} & \multicolumn{2}{c}{\textbf{Recall@1}}    \\
    & AV-Encoder & Pointer &2DP-Encoder &0.5  & 0.7 \\

    \midrule
    \multirow{1}{*}{\textbf{2DP-2MRC}}  & $\checkmark$ & $\checkmark$ & $\checkmark$ & \textbf{79.27}& \textbf{62.69}   \\
    \midrule
    \multirow{1}{*}{\textbf{1}}  & $\times$ & $\checkmark$ & $\checkmark$ & 74.09& 39.38   \\

   \midrule
    \multirow{1}{*}{\textbf{2}}  & $\checkmark$ & $\times$ & $\checkmark$ & 74.61 & 55.96  \\

   \midrule
    \multirow{1}{*}{\textbf{3}}  & $\checkmark$ & $\checkmark$ & $\times$ & 78.76 & 57.00  \\

    \bottomrule
  \end{tabular}
  
  }

\end{table}
\vspace{-3mm}


\begin{figure}[t]

\begin{minipage}[b]{1.0\linewidth}
  \centering
  \centerline{\includegraphics[width=8cm]{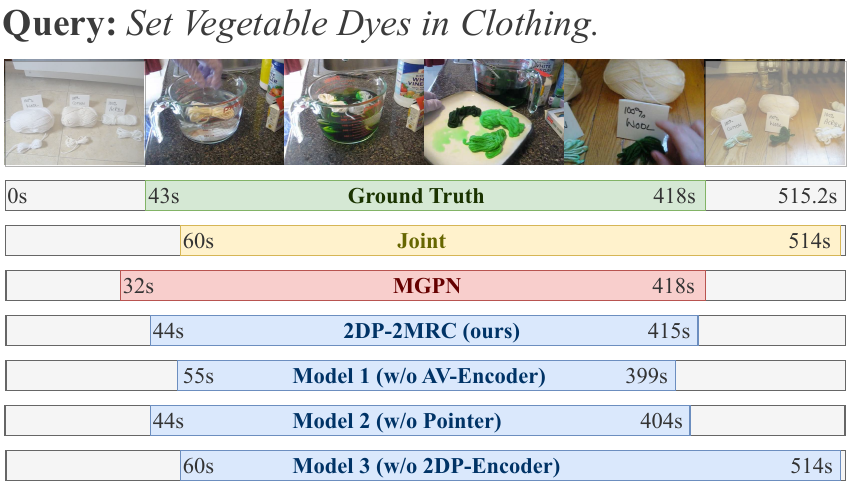}}
\end{minipage}
\vspace{-6mm}
\caption{A qualitative example of our 2DP-2MRC and ablation models evaluated on the HIREST dataset (with 128 video clips).}
\label{fig:heatmap}
\vspace{-6mm}
\end{figure}


\textbf{Qualitative Analysis}. To demonstrate the effectiveness of 2DP-2MRC, we further conduct qualitative analysis. As shown in Fig. \ref{fig:heatmap}, our 2DP-2MRC precisely retrieves moments most relevant to the language query. We also present the qualitative results of Model 1, 2, and 3 discussed in the ablation study. It can be observed that Model 1, lacking the coarse-grained feature encoder module, lacks overall attention to the video, resulting in relatively narrow retrieval moments. Similarly, Model 3, without the 2DP-Encoder module, struggles to capture the end moment boundary, especially when video segments between 418s and 514s are visually similar, a scenario also observed in the Joint model. Overall, our model learns the interaction between modalities in a coarse-to-fine manner, akin to human reading habits, prompting more accurate retrieval.

\vspace{-2mm}
\section{Conclusion and Future Work}
This paper proposes a novel \textbf{2-D}imensional \textbf{P}ointer-based \textbf{M}achine \textbf{R}eading \textbf{C}omprehension for \textbf{M}oment \textbf{R}etrieval \textbf{C}hoice (2DP-2MRC) model, which aligns with human reading comprehension habits. Our model achieves more accurate moment localization by designing AV-Encoder, Pointer, and 2DP-Encoder modules to interact with modalities at different granularities, akin to the strategy of initially reading the passage and question, focusing on the passage again, and integrating information to find the answer. Experimental results on the HIREST dataset demonstrate the effectiveness of our proposed 2DP-2MRC. Given that HIREST is a multi-task dataset, we plan to innovate on tasks such as moment segmentation and video captioning in future research.

\noindent
\textbf{Acknowledgements.} This work was supported in part by JST CREST Grant Number JPMJCR22D1, Japan, and a project, JPNP20006, commissioned by NEDO.

\bibliographystyle{IEEEtran}
\bibliography{mybib}

\end{document}